\documentclass[10pt,twocolumn,letterpaper]{article}

\usepackage[pagenumbers]{iccv} 

\AtBeginDocument{%

}

\usepackage{amsmath,amsfonts,bm}
\usepackage{stmaryrd}

\def\eqref#1{equation~\ref{#1}}

\def\1{\bm{1}}

\DeclareMathAlphabet{\mathsfit}{\encodingdefault}{\sfdefault}{m}{sl}
\SetMathAlphabet{\mathsfit}{bold}{\encodingdefault}{\sfdefault}{bx}{n}

\usepackage{xspace}
\newcommand{\methodname}{AdaHuman}
\newcommand{\mn}{\methodname\xspace}
\newcommand{\cmark}{\ding{51}}%
\newcommand{\xmark}{\ding{55}}%

\definecolor{iccvblue}{rgb}{0.21,0.49,0.74}
\usepackage[pagebackref,breaklinks,colorlinks,allcolors=iccvblue]{hyperref}
\usepackage{ulem}
\usepackage{pifont}
\usepackage{xcolor}

\def\confYear{2025}

\title{\methodname: Animatable Detailed 3D \mbox{Human} Generation with Compositional Multiview Diffusion}

\author{
Yangyi Huang \hspace{.5em} Ye Yuan \hspace{.5em} Xueting Li \hspace{.5em} Jan Kautz \hspace{.5em} Umar Iqbal \\[1mm]
NVIDIA \\
{ \url{https://nvlabs.github.io/AdaHuman}} \\
}

\begin{document}

\twocolumn[{%
\renewcommand\twocolumn[1][]{#1}%
\maketitle
 \begin{center}
     \vspace{-7mm}
     \centering
     \includegraphics[trim={1cm 0 0 0}, width=1.0\textwidth]{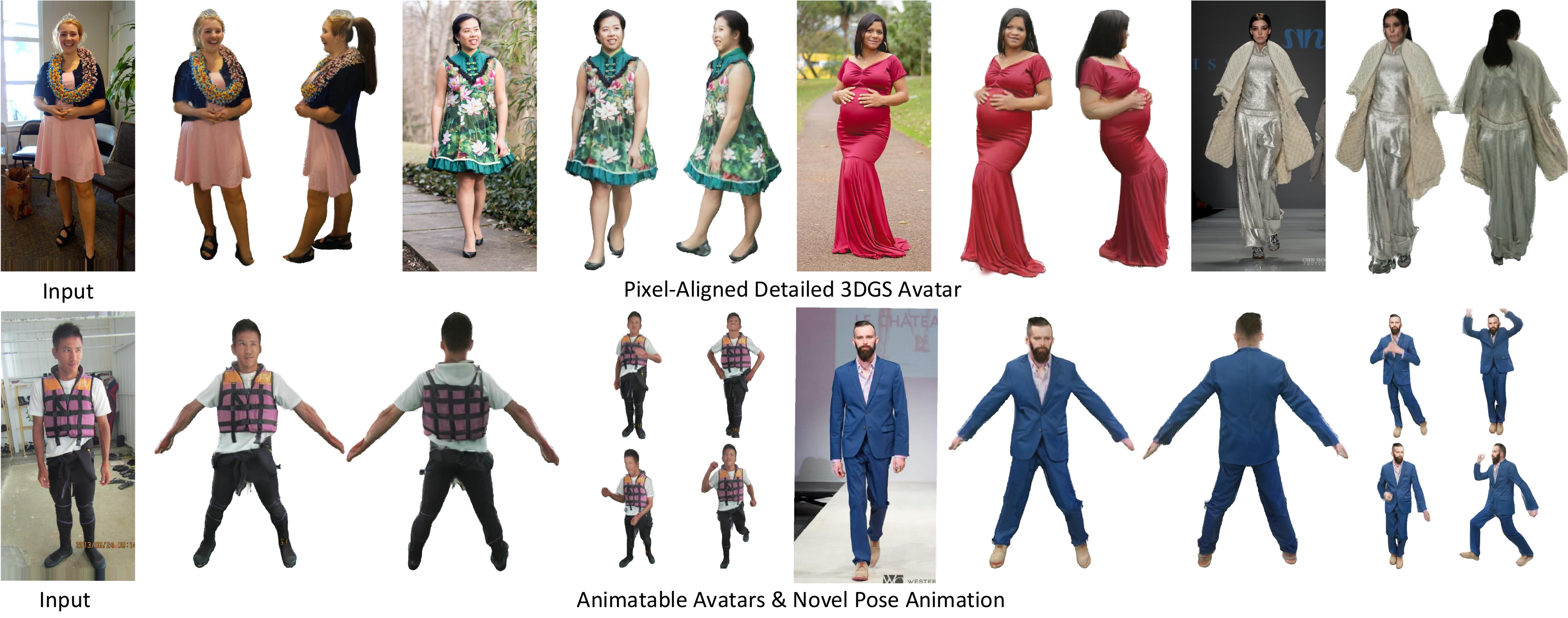}
     \vspace{-8mm}
     \captionof{figure}{Given a single input image, \mn reconstructs pixel-aligned a 3DGS avatar with detailed appearance. It can also generate the same avatar in novel poses, or in a standard animation-friendly A-pose to build an animatable avatar.}
     \vspace{0mm}
     \label{fig:teaser}
 \end{center}%
}]
\maketitle
\begin{abstract} 
     Existing methods for image-to-3D avatar generation struggle to produce highly detailed, animation-ready avatars suitable for real-world applications. We introduce \mn, a novel framework that generates high-fidelity animatable 3D avatars from a single in-the-wild image. \mn incorporates two key innovations: (1) A pose-conditioned 3D joint diffusion model that synthesizes consistent multi-view images in arbitrary poses alongside corresponding 3D Gaussian Splats (3DGS) reconstruction at each diffusion step; (2) A compositional 3DGS refinement module that enhances the details of local body parts through image-to-image refinement and seamlessly integrates them using a novel crop-aware camera ray map, producing a cohesive detailed 3D avatar. These components allow \mn to generate highly realistic standardized A-pose avatars with minimal self-occlusion, enabling rigging and animation with any input motion. Extensive evaluation on public benchmarks and in-the-wild images demonstrates that \mn significantly outperforms state-of-the-art methods in both avatar reconstruction and reposing. Code and models will be publicly available for research purposes.
\end{abstract}
\vspace{-10mm}
\section{Introduction}
\label{sec:intro}
Generating high-quality animatable 3D human avatars is crucial for numerous applications in gaming, animation, and virtual reality. Recent advances in diffusion-based image generation models have significantly accelerated research in this domain. Early approaches tackled this challenge using score distillation sampling (SDS)~\cite{poole2022dreamfusion}, where a 3D model is distilled from a diffusion-based image generation model~\cite{huang2024humannorm,huang2024tech}. While SDS-based methods offer flexibility and compatibility with various 3D representations, they suffer from oversaturation artifacts and slow generation speed, making them impractical for large-scale avatar creation. More recent methods have shifted towards multi-view generation and reconstruction pipelines~\cite{xue2024human3diffusion}, where diffusion models first synthesize multi-view images from text or image inputs, followed by a reconstruction phase that converts these images into a 3D avatar. This feed-forward approach improves both realism and generation speed. However, significant challenges remain. First, the avatars are typically generated in the same pose as the input image, leading to self-occlusion issues that complicate rigging and animation. Second, the resulting avatars often lack fine details and appear blurry, limiting their utility in real-world applications.

Motivated by these challenges, we introduce \mn, a new framework for generating animatable high-fidelity 3D human avatars from a single input image. At its core, \mn employs a pose-conditioned joint 3D diffusion model that seamlessly integrates multi-view image synthesis with 3D Gaussian Splats (3DGS)-based reconstruction during the diffusion process. By performing 3D reconstruction at each diffusion step, our approach ensures strong multi-view consistency across generated images, resulting in high-quality 3DGS avatars. A key advantage of our multi-view diffusion model is its ability to generate images in any arbitrary pose by simply conditioning on the desired pose. To enable animation, we leverage this capability to generate the 3DGS avatar in a standard A-pose, which minimizes self-occlusion, inpaints missing details, and naturally facilitates rigging and animation. Notably, our method achieves this without requiring training images in such standard poses.

To enhance the fidelity and detail of the generated avatars, \mn further introduces a compositional 3DGS refinement module. This module first renders zoomed-in views of local body parts (e.g., head, upper body, lower body) from the initial 3DGS avatar. These local views then undergo an image-to-image refinement process using our multi-view diffusion model to improve detail and resolution. Using these refined local views, we propose a novel approach that seamlessly integrates the local views and global full-body views to produce a highly detailed holistic 3D avatar. This is enabled through two innovations: (1) a crop-aware camera ray map that establishes precise correspondences between 3D locations in local and global views, and (2) a visibility-aware composition scheme that intelligently merges partial 3DGS reconstructions based on view coverage and visibility salience. Our approach effectively prevents floating artifacts while preserving fine details and coherency, resulting in high-quality 3D avatars with enhanced local and global consistency.

In summary, our main contributions are as follows: (1) We introduce a new image-to-avatar framework leveraging pose-conditioned 3D joint diffusion, enabling both avatar reconstruction and reposing for seamless rigging and animation. (2) We develop an innovative compositional 3DGS refinement approach that produces highly detailed and globally consistent avatars using a crop-aware camera ray map and a visibility-aware composition scheme. (3) Through comprehensive evaluation on public benchmarks and challenging in-the-wild images, we demonstrate that our method substantially outperforms state-of-the-art approaches in both avatar reconstruction and reposing tasks.

\section{Related Work}
\label{sec:related-work}

\noindent\textbf{3D Avatar Reconstruction.} 
Early methods for monocular RGB-based 3D avatar reconstruction typically rely on the SMPL~\cite{loper2015smpl, pavlakos2019smplx} model, predicting per-vertex offsets to capture clothing and hair details, but are limited by SMPL's fixed topology. 
Consequently, recent approaches adopt implicit representations allowing arbitrary topologies~\cite{saito2019pifu, zhi2020texmesh, saito2020pifuhd, huang2020arch, xiu2023icon, xiu2023econ, yang2021s3, alldieck2022phorhum, huang2024tech, zhang2023sifu, ho2024sith}, however, they depend heavily on extensive 3D training data. Moreover, they struggle with occlusion handling in complex poses making it difficult to animate the reconstructed avatars. 
Methods enabling animation through pose canonicalization usually require ground-truth standard-pose meshes or rigged avatars~\cite{huang2020arch,he2021arch++,peng2024charactergen}. In contrast, our method generalizes reposing from diverse multiview video data, directly generating avatars in arbitrary poses without relying on standard-pose training data.

\vspace{1mm}
\noindent\textbf{3D Avatars Generation via 2D Foundation Model.} 
Advances in 2D diffusion models~\cite{Rombach2022StableDiffusion, dalle2} have driven significant progress in 3D avatar generation~\cite{poole2022dreamfusion, lin2022magic3d, richardson2023texture, chen2023fantasia3d, wang2023prolificdreamer, dreamgaussian, shi2023mvdream, cao2023dreamavatar, kolotouros2023dreamhuman, jiang2023avatarcraft, zhang2023getavatar, huang2023dreamwaltz, zhang2023avatarverse, huang2024humannorm, liao2024tada, yuan2024gavatar}. These methods adopt the Score Distillation Sampling (SDS) technique to extract 3D knowledge from these models. SDS-based methods, however, suffer from unrealistic outputs and slow iterative optimization, resulting in lower quality avatars and prohibitively long run time for wider adoption. 

\vspace{1mm}
\noindent\textbf{Joint Diffusion and Reconstruction.}
Recent methods combine diffusion models with reconstruction networks to improve efficiency and quality for 3D avatar generation~\cite{liu2023one2345, liu2023one2345pp, shi2023zero123++, long2023wonder3d, xu2024instantmesh, Xu2023DMV3D, zou2023triplanegaussian}. 
Zero123~\cite{Liu2023zero123} and its variants~\cite{liu2023one2345, liu2023one2345pp, shi2023zero123++, long2023wonder3d}  generate consistent multi-view images that facilitate accurate 3D avatar reconstruction.  More recent works~\cite{hong2023lrm, tang2025lgm, tochilkin2024triposr, xu2024grm} predict implicit 3D representations directly from multi-view images, enabled by advancements in implicit representations such as triplanes~\cite{chan2022eg3d} and Gaussian Splats~\cite{kerbl20233d}. Similar strategies have been applied to human avatars~\cite{albahar2023humansgd, huang2024tech, sengupta2024diffhuman, kolotouros2024avatarpopup}, though they remain limited by the quality of generated views. Recently, Xue et al.~\cite{xue2024human3diffusion} proposed a method that jointly trains diffusion and reconstruction models in an end-to-end manner, allowing for mutual enhancement.

Our method follows this direction but introduces key innovations: (1) a pose-conditioned multi-view joint diffusion model that synthesizes avatars in arbitrary poses to handle occlusions and facilitate animation; (2) a compositional 3DGS refinement strategy integrating global and local views via a crop-aware camera ray embedding, significantly enhancing avatar detail and coherence. 

\noindent\textbf{Concurrent works.} Some of the latest research, IDOL~\cite{zhuang2024idolinstantphotorealistic3d} and LHM~\cite{qiu2025LHM} also trying reconstruct high-resolution 3DGS avatars with large scale training data. While they develops feed-forward models for efficiency, we build our model based on diffusion models to utilize the strong generative priors.
\section{Approach}
\label{sec:approach}

\begin{figure*}[h]
\begin{center}
 \centering
 \includegraphics[width=1.0\textwidth]{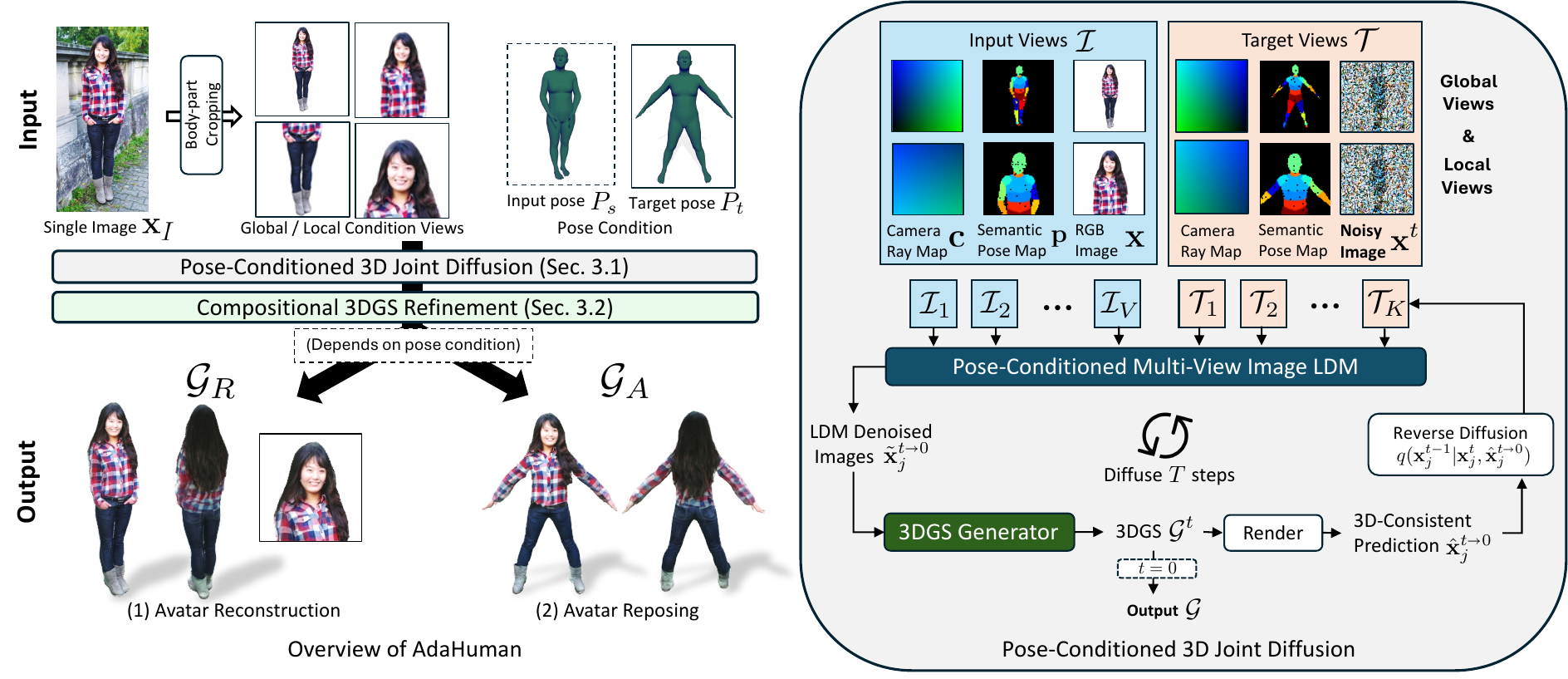}
 \vspace{-2em} 
 \caption{\textbf{Method Overview.} Left: Given an RGB image of an unseen person as input, \mn could (1) reconstruct a high-fidelity pixel-aligned 3D Gaussian Splat (3DGS) avatar, as well as (2) generate an reposed 3DGS avatar with a target pose condition, enable building animatable avatar in a standard A-pose. Right: A pose-conditioned joint 3D diffusion process is utilized to generate global or local 3DGS reconstruction or reposing results. This process ensures 3D consistency of the reconstruction by utilizing generated 3DGS results in each reverse diffusion process of multi-view avatar images.}
 \label{fig:approach-overview}
 \vspace{-2em} 
\end{center}
\end{figure*}

\textit{Problem Specification}. As illustrated in \cref{fig:approach-overview}, given a full-body input image $\mathbf{x}_I$ depicting a person, \mn aims to build a 3D avatar that supports two key functionalities:
(1) \textbf{Avatar Reconstruction}: Without requiring any additional inputs, \mn reconstructs an avatar $\mathcal{G}_R$ represented by 3D Gaussian Splats (3DGS)~\cite{kerbl20233d} that precisely matches the pose of the input image, enabling high-fidelity novel view synthesis;
(2) \textbf{Avatar Synthesis}: Using an estimated input pose $P_s$ and an arbitrary target 3D pose $P_t$, \mn generates a reposed 3DGS avatar $\mathcal{G}_{P_t}$ in the target pose while faithfully preserving the person's appearance and identity. This capability enables pose canonicalization, where we generate a standardized A-posed avatar $\mathcal{G}_A$ that minimizes self-occlusion. The canonicalized avatar can then be rigged automatically for animation and used to render temporally coherent 4D videos with high visual quality.

Our method consists of two key modules that enable the generation of detailed and animatable avatars: (1) Pose-Conditioned 3D Joint Diffusion (\cref{sec:pose-conditioned-3d-joint-diffusion}), which generates multiview images and the corresponding 3DGS avatar of the person in arbitrary poses by interleaving image synthesis and 3D reconstruction inside the diffusion process; (2) Compositional 3DGS Refinement (\cref{sec:compose-3dgs}), which enhances the visual quality by first refining local body part renderings at high resolution and then seamlessly composing them into a holistic detailed avatar.

\subsection{Pose-Conditioned 3D Joint Diffusion}
\label{sec:pose-conditioned-3d-joint-diffusion}
As shown in \cref{fig:approach-overview}, given a full-body input image $\mathbf{x}_I$, we first generate local view images of different body parts (\eg, head, upper body, and lower body). These local views, along with the input, form our input views $\mathcal{I}_{i=1}^V$, which are fed to the 3D joint diffusion module as in \cref{fig:approach-overview}\,(right). The module then synthesizes images of the target views $\mathcal{T}_{j=1}^K$ which look at the full-body and local body parts of the person from different viewpoints than the input.
Combining both full-body and local perspectives enables our method to achieve detailed and globally consistent generation of multi-view images and their corresponding 3DGS avatar.

Each input view is represented by a tuple $\mathcal{I}_i = \{\mathbf{x}_i, \mathbf{p}_i, \mathbf{c}_i\}$, consisting of an RGB image $\mathbf{x}_i$, an \textit{optional} pose condition $\mathbf{p}_i$, and camera parameters $\mathbf{c}_i$.
The pose condition $\mathbf{p}_i$ takes the form of a 2D semantic pose map derived from the 3D input pose $\theta$, created by rendering the semantic segmentation of the SMPL model~\cite{loper2015smpl} from the camera's perspective. 
The camera parameters $\mathbf{c}_i$ are encoded into a camera ray map using sinusoidal embeddings of the camera rays' origins and directions.
Similarly, each target view is defined by $\mathcal{T}_j = \{\mathbf{x}_j^t, \mathbf{p}_j, \mathbf{c}_j\}$, where $\mathbf{x}_j^t$ represents the noisy target RGB image at diffusion step $t$, $\mathbf{p}_j$ is the optional pose condition, and $\mathbf{c}_j$ encodes the target view's camera parameters. The primary objective of our pose-conditioned 3D joint diffusion is to model the conditional denoising distribution of the target RGB images $\{\mathbf{x}_j^{t-1}\}_{j=1}^K$:
\begin{equation}
\label{equ:mv-diffusion}
    p(\{\mathbf{x}_j^{t-1}\}_{j=1}^K|\{\mathbf{p}_j, \mathbf{c}_j\}_{j=1}^K, \{\mathbf{x}_i, \mathbf{p}_i, \mathbf{c}_i\}_{i=1}^V, t)\,,
\end{equation}
where we assume $V$ input views and $K$ target views. Inspired by recent work~\cite{gao2024cat3d}, we employ a multi-view image latent diffusion model (LDM) to model the denoising distribution. Specifically, we modify the U-Net architecture of a single-image LDM by replacing the 2D self-attention layers with 3D attention layers. The 2D pose semantic map $\mathbf{p}_i$ and camera ray map $\mathbf{c}_i$ are concatenated with the RGB images as additional conditions before being fed to the U-Net.

To enhance the 3D consistency of the generated multi-view images and produce the underlying 3DGS avatar, we incorporate a 3DGS generator $\mathbf{G}$~\cite{tang2025lgm} into the denoising diffusion process. 
Following~\cite{xue2024human3diffusion}, at each denoising step $t$, we generate a 3DGS avatar $\mathcal{G}^t$ from the image predictions:
\begin{equation}
\mathcal{G}^t = \mathbf{G}(\{\mathbf{x}_j^{t\shortrightarrow 0},\mathbf{x}_j^{t},\mathbf{p}_j, \mathbf{c}_j\}_{j=1}^K, \{\mathbf{x}_i, \mathbf{p}_i, \mathbf{c}_i\}_{i=1}^V, t)\,, 
\label{equ:joint-diffusion}
\end{equation}
where $\mathbf{x}_j^{t\shortrightarrow 0}$ represents the ``clean'' target image obtained through one-step denoising by the LDM at diffusion step $t$. Once $\mathcal{G}^t$ is obtained, we render it under the target views to generate new \textit{3D-consistent} clean target images $\hat{\mathbf{x}}_j^{t\shortrightarrow 0}$. Using these 3D-consistent images, we then sample the noisy images $\mathbf{x}_j^{t-1}$ for the next diffusion step according to:
$\mathbf{x}_j^{t-1} \sim q(\mathbf{x}_j^{t-1}|\mathbf{x}_j^{t},\hat{\mathbf{x}}_j^{t\shortrightarrow 0})$,
where $q$ denotes the reverse diffusion process~\cite{ddim}. The final output of our pose-conditioned 3D joint diffusion model is the 3DGS avatar $\mathcal{G}^0$ produced at the end of the diffusion process.

Unlike previous works~\cite{gao2024cat3d, xue2024human3diffusion}, our approach incorporates pose conditioning to enable pose-conditioned multi-view image synthesis. This key enhancement empowers our model to not only reconstruct pixel-aligned 3DGS avatars but also generate reposed avatars that are well-suited for animation and other applications. This capability is particularly valuable since subjects in input images often exhibit severe self-occlusion, which makes rigging in the original body pose challenging and suboptimal. Through pose-conditioned multi-view image synthesis, our method can transition the avatar into a rigging-friendly pose while simultaneously recovering geometry and appearance details that were previously occluded.

\vspace{2mm}
\noindent\textbf{View Selection and Model Training.}
During training,
we first randomly select either the full body or a local body parts from upper body, lower body, or head. For the selected body part, we choose an input view from a training video frame. The key distinction between reconstruction and reposing lies in the selection of target views: for reconstruction, we select three canonical target views (separated by 90° azimuth angles) of the body part from the \textit{same} frame as the input view; for reposing, we select four canonical target views from a \textit{different} frame showing the subject in a different pose, where the additional target view coincides with the input view to account for the pose difference.

We jointly train the multi-view image LDM and the 3DGS generator $\mathbf{G}$ using multi-view video data from MVHumanNet~\cite{xiong2024mvhumannet} and image renderings from CustomHuman~\cite{ho2023custom}. To leverage powerful generative priors learned from large-scale datasets, both models are initialized from official pretrained weights~\cite{Rombach2022StableDiffusion, tang2025lgm}. We first train the model for avatar reconstruction for 30k steps and then fine-tune the model for reposing for 10k steps. Camera ray embeddings are computed relative to the input view. The LDM is supervised using MSE loss between predicted and ground truth image latents, while the 3DGS generator $\mathbf{G}$ is supervised following~\cite{xue2024human3diffusion} using MSE, LPIPS rendering losses, and surface regularization loss. In addition to the target views, we sample 12 additional views to provide dense supervision to the 3DGS generator. Additional implementation details are provided in 
the appedix.

\begin{figure*}[h]
\begin{center}
 \centering
 \includegraphics[width=1.0\linewidth]{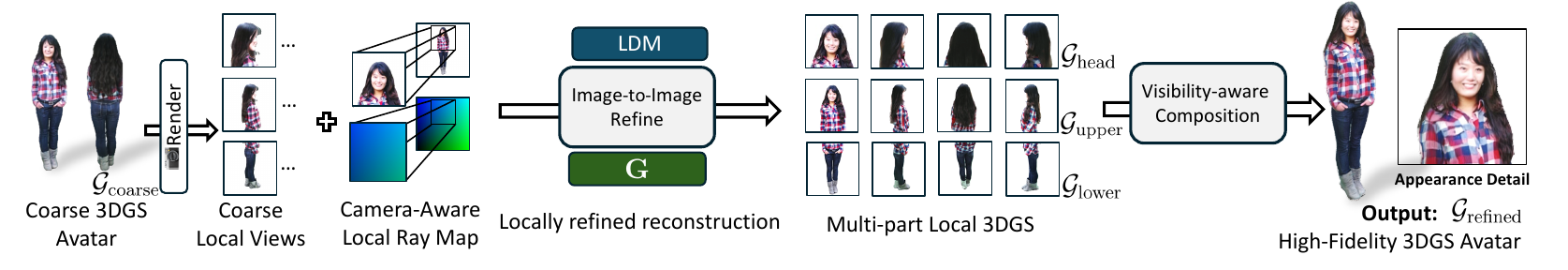}
 \vspace{-8mm}
 \caption{\textbf{Compositional 3DGS Refinement.} Given the coarse 3DGS reconstruction $\mathcal{G}_\mathrm{coarse}$ as input, we render initial coarse views, and refine them with image-to-image editing for enhancing local 3DGS $\mathcal{G}_\mathrm{upper},\mathcal{G}_\mathrm{lower},\mathcal{G}_\mathrm{head}$. Finally, a refined holistic 3DGS avatar $\mathcal{G}_\mathrm{refined}$ is generated from these results by our proposed visibility-aware 3DGS Composition.} 
 \label{fig:compose}
 \vspace{-2em}
\end{center}
\end{figure*}

\subsection{Compositional 3DGS Refinement}
\label{sec:compose-3dgs}

Recent feed-forward 3D reconstruction models~\cite{hong2023lrm,tang2025lgm} have demonstrated promising results in generating 3D models of general objects from sparse-view images. However, these approaches are constrained by their networks' fixed output resolution (e.g., $256{\times}256$ 3D Gaussians in LGM~\cite{tang2025lgm}), limiting their ability to capture the fine-grained details essential for realistic human avatar reconstructions. To address this limitation, we introduce a new compositional 3DGS refinement module, as illustrated in \cref{fig:compose}. The module leverages an image-to-image local body refinement scheme as well as a novel crop-aware camera ray map to enable detailed and coherent reconstructions of individual local body parts. During inference, it takes the coarse 3DGS avatar $\mathcal{G}_\text{coarse}$ from the 3D joint diffusion module as input and refines it to produce a detailed 3DGS avatar $\mathcal{G}_\text{refined}$.

\vspace{2mm}
\noindent\textbf{Local body part refinement.} To achieve enhanced details for local body parts, we begin by rendering $N_v{=}4$ 90-degree separated canonical views (front, left, back, and right) for each of $N_b{=}3$ local body parts (head, upper body, and lower body) of the coarse avatar $\mathcal{G}_\text{coarse}$. Each local view is produced using a crop-view camera that zooms into the local body region inside the original global view (by manipulating the camera intrinsics). This zoom-in region is computed using the 2D body joints and segmentation masks. We then employ our multi-view LDM introduced in Section~\ref{sec:pose-conditioned-3d-joint-diffusion} to refine the local renderings via an image-to-image editing process similar to SDEdit~\cite{meng2021sdedit}, significantly enhancing their detail. This approach enables the high-fidelity generation of local body parts. To properly handle the modified camera perspective for these local views, we provide the LDM with a specialized cropped version of the camera ray map, which we detail in the following section.

\vspace{2mm}
\noindent\textbf{Crop-aware local ray map.}
A key challenge in the refinement process is effectively combining the $N_v{\times}N_b$ refined local view images and $N_v$ global full-body view images into a holistic 3DGS avatar. The 3DGS generator in~\cite{tang2025lgm} uses four fixed canonical camera views as inputs to generate a global 3DGS in unit space, but this fixed camera setup does not naturally accommodate additional local views.

To address this challenge, we propose a simple yet effective solution: a crop-aware local ray map that establishes correspondences between the 3D coordinates of local and global views. This approach extends \cite{tang2025lgm} by incorporating additional local views as inputs, enabling high-resolution reconstruction of local body parts with fine details. Specifically, for a pixel at coordinates $(u, v)$ in a local view image of size $(H, W)$, where the local view is obtained by cropping a box region $(x_{tl}, y_{tl}, x_{br}, y_{br})$ from the global view, we map its coordinates back to the global view using:
\begin{equation}
    (i, j) = \left(x_{tl} + \frac{(x_{br} - x_{tl}) \cdot u}{W},
     y_{tl} + \frac{(y_{br} - y_{tl}) \cdot v}{H}\right).
\end{equation}
Using these mapped coordinates, we compute the camera ray embedding for the local view pixel using the 3DGS generator's global camera ray map equation:
\begin{equation}
    \mathcal{R}(i, j) = (\mathbf{o}(i, j), \mathbf{o}(i, j) \times \mathbf{d}(i, j))
    \label{equ:lgm-ray-embed}
\end{equation}
where $\mathbf{o}$ and $\mathbf{d}$ represent the origin and direction of the camera rays based on the camera extrinsics. The crop-aware local ray map is utilized during both training and inference to help the 3DGS generator establish correspondences between the 3D locations in local and global views. Using the crop-aware local ray map, we can directly use the 3DGS generator $\mathbf{G}$ to map refined local views to 3DGS in the global avatar space. In the following, we will describe a strategy to combine the 3DGS produced by the local and global views into a holistic 3DGS avatar $\mathcal{G}_\text{refined}$.

\vspace{2mm}
\noindent\textbf{Visibility-aware 3DGS Composition.}
As we will show in \autoref{fig:ablation}, naively combining these partial 3DGS leads to floating artifacts and degraded appearance details. To address this challenge, we introduce a visibility-aware 3DGS composition scheme that intelligently merges the parts into a coherent, high-quality avatar. Our approach employs two key criteria to determine which 3D Gaussians to preserve during composition: (1) \textit{View Coverage} quantifies how many input views capture each 3D Gaussian point within their field of view, and (2) \textit{Visibility Salience} measures the gradient magnitude of the alpha channel across all rendered input views. Intuitively, Gaussians with low view coverage lack multi-view consensus and are likely unreliable, while those with low visibility salience contribute minimally to the final appearance and likely represent noise. 
Specifically, given globally or locally reconstructed body part 3DGS $\mathcal{G}_\mathrm{p}$ and the canonical views for each body part $\mathcal{I}_\mathrm{p}^j$, where $p\in\{\mathrm{full}, \mathrm{upper}, \mathrm{lower}, \mathrm{head}\}$ and $j= 0\dots3$, we evaluate each splat $\mathcal{G}_\mathrm{p}^i$ as follows:

First, we calculate the number of covered input views of the splat in different local parts $n_c(\mathcal{G}_\mathrm{p_1}^i, \mathcal{I}_\mathrm{p_2})$. A splat is considered reliable if it is covered by more than 2 input views in its own body part ( or 3 views if it is generatd by the head part. If the splat is also well-covered by input views of another more detailed body part (e.g., head is more detailed than upper-body), it is deemed redundant and removed.

Second, we assess visibility salience using rendering gradients. If a splat has higher visibility in the input views of another body parts with similar level of detail (e.g., between upper and lower body), it is likely redundant and should be dropped to avoid conflicts or redundancy.

This approach ensures efficient composition while maintaining visual fidelity, focusing on the most reliable and visually significant splats.

\section{Experiments}
\label{sec:experiments}

In order to comprehensively evaluate the performance of \mn, we conduct experiments on avatar reconstruction and avatar reposing tasks, comparing our method with state-of-the-art (SOTA) approaches both quantitatively and qualitatively. Additionally, we perform a user study to assess the perceptual quality of the generated avatars.

\begin{figure*}[h]
\begin{center}
 \centering
 \includegraphics[width=1.0\linewidth]{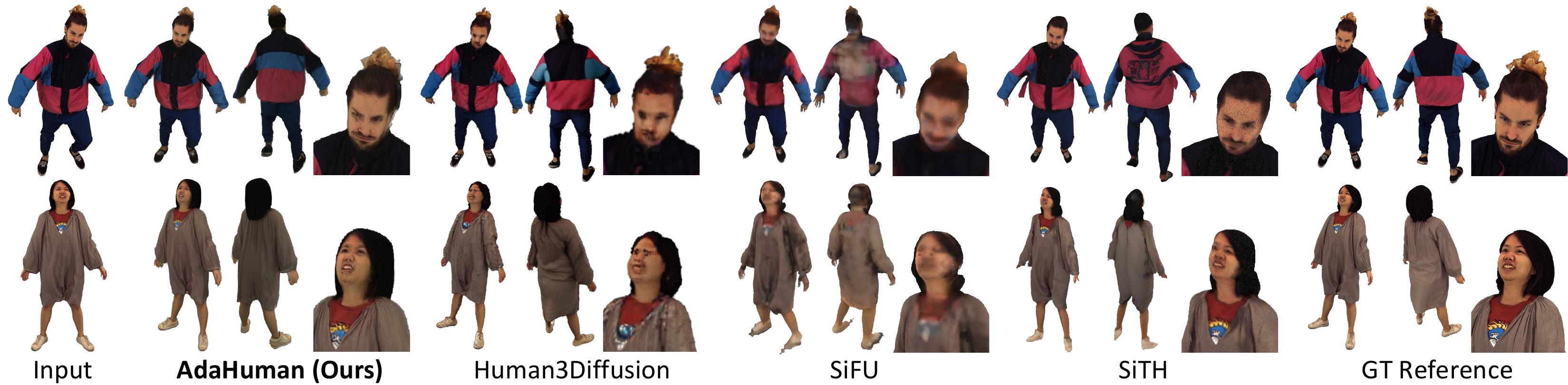}
    \vspace{-1.8em}
 \caption{\textbf{Qualitative comparison on vatar reconstruction task on CustomHumans\cite{ho2023custom} dataset.}} 
    \vspace{-2.0em}
 \label{fig:exp-comp-nvs-ch}
\end{center}
\end{figure*}

\begin{figure*}[h]
\begin{center}
 \centering
 \includegraphics[width=1.0\linewidth]{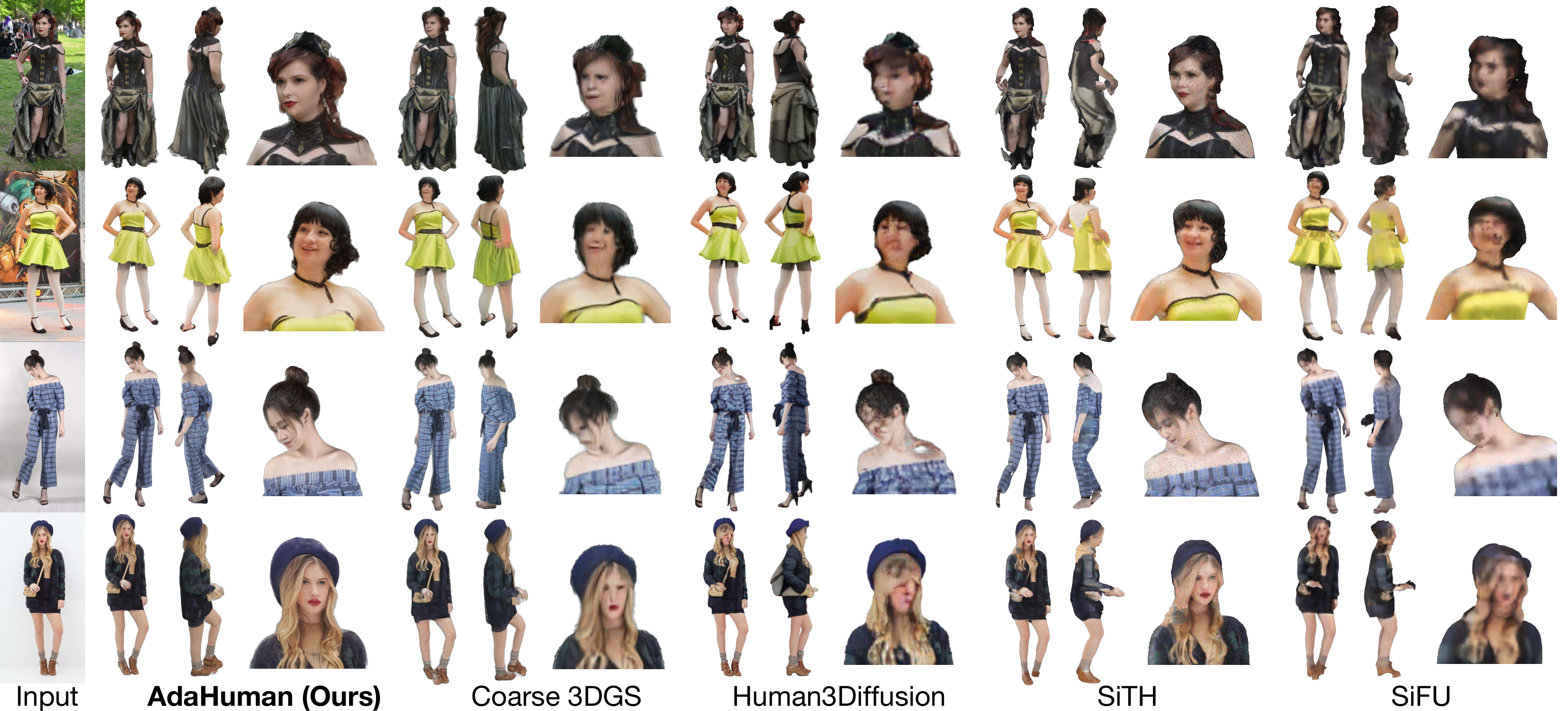}
 \vspace{-2em}
 \caption{\textbf{Comparison on in-the-wild images.} AdaHuman generalizes well to images with diverse appearances, body shapes, and clothing styles, while SIFU\cite{zhang2023sifu} and SiTH\cite{ho2024sith} fail on loose and complex clothing, and Human3Diffusion~\cite{xue2024human3diffusion} fail to preserve appearance details. Coarse 3DGS is an ablation variant of AdaHuman without compositional 3DGS refinement, which fails to capture fine avatar details.} 
 \vspace{-1.8em}
 \label{fig:exp-comp-shhq}
\end{center}
\end{figure*}

\begin{table*}[ht]
    \centering
    \scriptsize
    \setlength{\tabcolsep}{4.0pt}
    \resizebox{\linewidth}{!}{
    \begin{tabular}{@{}lcc|cc|cc|cc|cc|cc|cc|cc}
        \textbf{Model} & \multicolumn{2}{c|}{\textbf{PSNR$\uparrow$}} & \multicolumn{2}{c|}{\textbf{SSIM$\uparrow$}} & \multicolumn{2}{c|}{\textbf{LPIPS$\downarrow$}} & \multicolumn{2}{c|}{\textbf{FID $\downarrow$}} & \multicolumn{2}{c|}{\textbf{CD}(cm)$\downarrow$} & \multicolumn{2}{c|}{\textbf{F-score}$\uparrow$} & \multicolumn{2}{c}{\textbf{Normal}$\uparrow$}\\ 
        & CH & Sizer & CH & Sizer & CH & Sizer & CH & Sizer & CH & Sizer & CH & Sizer & CH & Sizer \\ \hline
        LGM~\cite{tang2025lgm} & 18.99 & 17.58 & 0.8445 & 0.8909 & 0.1664 & 0.1188 & 122.3 & 124.20 & 2.175 & 1.832 & 0.3941 & 0.4897 & 0.6431 & 0.6451 \\
        SiTH~\cite{ho2024sith} & 20.77 & 20.67 & 0.8727 & 0.9219 & 0.1277 & 0.0883 & 42.9 & 37.11 & 1.389 & 1.229 & 0.4701 & 0.5688 & \textbf{0.7978} &  \textbf{0.7915} \\ 
        SIFU~\cite{zhang2023sifu} $\dagger$ & 20.59 &  20.56 &  0.8853 & 0.9196 & 0.1359& 0.0987 & 92.6 & 101.79 & 2.009 & 1.560 & 0.3438 & 0.4787 & 0.7539 & 0.7768 \\            
        Human3Diffusion~\cite{xue2024human3diffusion}& 21.08 & 19.50 & 0.8728 & 0.9211 & 0.1364 & 0.0953 & 35.3 & 20.69 & 1.230 & 1.174 & 0.5324 & \textbf{0.6336} & 0.7338 & 0.7389 \\
        \mn (Ours) & \textbf{21.46} & \textbf{21.42} & \textbf{0.8925} & \textbf{0.9258} & \textbf{0.1087} & \textbf{0.0856} & \textbf{27.3} & \textbf{19.15}& \textbf{0.962} & \textbf{1.135} & \textbf{0.6083} & 0.6075 & 0.7597 & 0.7477 \
    \end{tabular}
    }
    \vspace{-1.0em}
    \caption{\textbf{Quantitative comparison on avatar reconstruction task.} On CustomHumans~(CH)~\cite{ho2023custom} and Sizer~\cite{tiwari2020sizer} datasets, \mn surpasses all baselines on rendering quality metrics (PSNR, SSIM, LPIPS and FID ), and also achieves best the shape reconstruction metrics (CD, F-score), except getting slightly lower F-score with Human3Diffusion~\cite{xue2024human3diffusion} on the Sizer dataset. However, since we borrow the same normal estimation method from \cite{xue2024human3diffusion}, \mn got similar performance on Normal Consistency. The \textbf{best} scores are highlighted. $\dagger$: not using SIFU's text-guided texture refinement since prompts are unavailable.}
    \vspace{-1.0em}
    \label{tab:nvs-customhumans}
\end{table*}

\vspace{2mm}
\noindent\textbf{Datasets.} Unlike most existing 3D avatar reconstruction methods that rely on 3D human mesh data for training, \mn leverages multi-camera video data from MVHumanNet~\cite{xiong2024mvhumannet}, which captures 3D appearances of humans in real-world settings and diverse poses. We sample 6,209 unique subjects for training and 50 unseen subjects for evaluating the novel pose synthesis task. Additionally, we mixed the training data with multiview images rendered from 589 human meshes in the CustomHumans~\cite{ho2023custom} dataset with a more diverse camera distribution to improve generalizability. 50 testing subjects from the CustomHumans~\cite{ho2023custom} dataset and 97 subjects from Sizer~\cite{tiwari2020sizer} dataset are used to quantitatively compare our method against SOTA approaches. To further assess visual quality, we use 53 in-the-wild human images from the SHHQ~\cite{fu2022styleganhuman} dataset to conduct a user study on perceptual quality.

\vspace{2mm}
\noindent\textbf{Runtime.}
Our whole pipeline takes around 70s  for inference on a NVIDIA A100 GPU.

\subsection{Avatar Reconstruction}

For novel view synthesis, we compare \mn with SOTA mesh reconstruction methods (SiTH~\cite{ho2024sith} and SIFU~\cite{zhang2023sifu}) and 3DGS-based methods (LGM~\cite{tang2025lgm} and Human3Diffusion~\cite{xue2024human3diffusion}) on the CustomHumans dataset~\cite{ho2023custom}.
For each test subject, we use a frontal camera view as the input image and render 20 novel views ($1024{\times}1024$) by rotating around the body. We follow~\cite{xue2024human3diffusion} to extract mesh from 3DGS results, and evaluate 3D reconstruction quality using Chamfer Distance~(CD), Normal Consistency~(NC) and F1 score. We evaluate rendering quality using PSNR, SSIM, and LPIPS scores for all novel views and the frontal view. FID scores are assessed to measure the perceptual quality of the avatars.
We provide qualitative and quantitative comparisons of \mn against SOTA methods in \cref{fig:exp-comp-nvs-ch} and \cref{tab:nvs-customhumans}. Our method generates significantly higher-quality avatars, with a better performance on all of image quality metrics, while keeping a comparable performance in the 3D reconstruction metrics. 

\subsection{Perceptual Study}

\begin{table}[ht]
    \centering
    \setlength{\tabcolsep}{1.0pt}
    \resizebox{\linewidth}{!}{
    \begin{tabular}{l|cccc}
         Baseline methods & SiTH\cite{ho2024sith} & SIFU\cite{zhang2023sifu} & H3D\cite{xue2024human3diffusion} & Coarse 3DGS\\
         \hline
         Preference of \mn ($\%$) & 88.3 & 99.2 & 79.7 & 93.8 
    \end{tabular}
    }
    \vspace{-1.0em}
    \caption{\textbf{User preference of \mn.} Our method achieves substantially higher preference against all baseline methods.}
    \vspace{-1.0em}
    \label{tab:user-study}
\end{table}

\begin{figure*}[h]
\begin{center}
 \centering
 \includegraphics[width=1.0\linewidth]{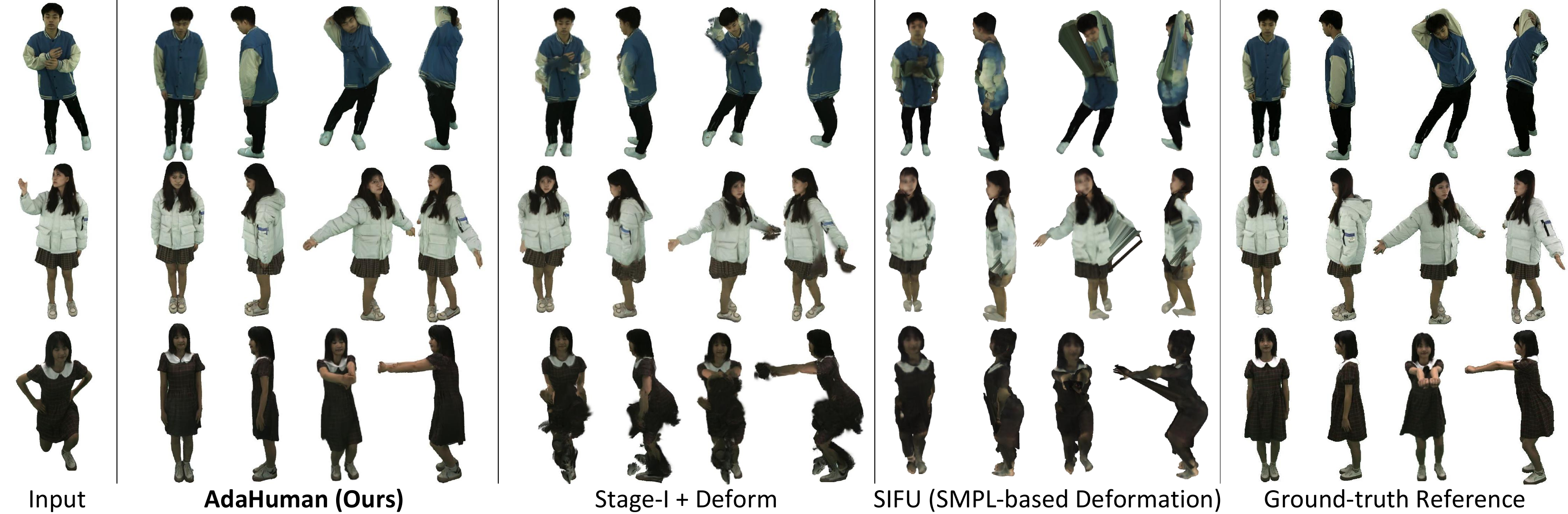}
 \vspace{-1.5em}
 \caption{\textbf{Qualitative comparison on novel pose synthesis task.}} 
 \vspace{-2em}
 \label{fig:exp-comp-nps}
\end{center}
\end{figure*}

\begin{figure*}[h]
\begin{center}
 \centering
 
 \vspace{-1.5em}
 \includegraphics[width=1.0\linewidth]{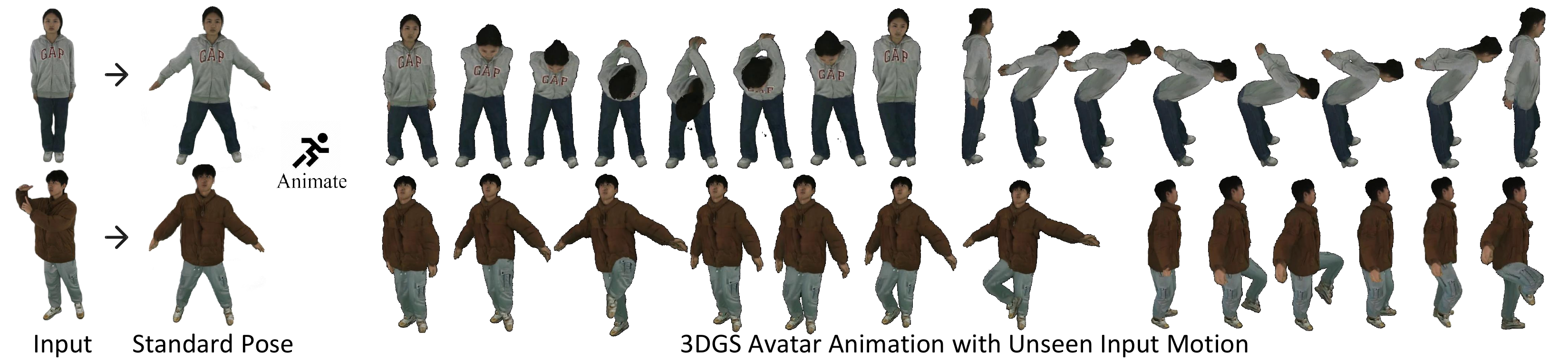}
 \vspace{-2em}
 \caption{AdaHuman generates animation-ready avatar in a standard pose, which can be animated with unseen input motion.}
 \vspace{-2em}
 \label{fig:exp-animation}
\end{center}
\end{figure*}

To fully evaluate the perceptual quality and generalizability of our method, we conducted a user study on 53 in-the-wild images from the SHHQ~\cite{fu2022styleganhuman} dataset. We compared \mn with SiTH~\cite{ho2024sith}, SIFU~\cite{zhang2023sifu}, Human3Diffusion~\cite{xue2024human3diffusion}, and an ablation of \mn using the coarse 3DGS avatar without compositional 3DGS refinement. Each survey consisted of 40 pairs of generated avatars, and 28 participants were asked to select the avatar with better overall quality.

As shown in \cref{tab:user-study}, \mn was preferred by a significant margin over other methods. \cref{fig:exp-comp-shhq} demonstrates that SIFU~\cite{zhang2023sifu} and SiTH~\cite{ho2024sith} often produce lower texture quality for side views and struggle to recover accurate geometry, likely due to the limitations of template-based mesh reconstruction. Our method generates avatars with substantially higher quality and generalizes well across diverse appearances, clothing styles, and body poses. Compared to Human3Diffusion~\cite{xue2024human3diffusion}, which fails to capture fine appearance details, our method recovers significantly better details thanks to our local refinement approach. More results on in-the-wild images are provided on the website.%

\subsection{Avatar Reposing and Animation}

\begin{table}[t]
    \centering
    \scriptsize
    
    \setlength{\tabcolsep}{5pt}
    \resizebox{0.9\linewidth}{!}{
    \begin{tabular}{l|cccc}
         Method & PSNR$\uparrow$ & SSIM$\uparrow$ & LPIPS $\downarrow$\\
         \hline
         SiTH & 21.21 & 0.8742 & 0.1261 \\
         SIFU & 21.27 & 0.8722 & 0.1244 \\
         \mn $\mathcal{G}_R $+ deform & 23.01 & 0.8825 & 0.1100\\
         \mn $\mathcal{G}_{P_t}$ (Ours) & \textbf{24.64} & \textbf{0.9046} & \textbf{0.0863}
    \end{tabular}
    }
    \vspace{-1.0em}
    \caption{\textbf{Comparison on novel pose synthesis task.} Our model achieves the best rendering similarity (PSNR, SSIM, LPIPS), showcasing the ability of our pose-conditioned model to generalize to diverse input and target poses.}
    \vspace{-2.0em}
    \label{tab:nps-mvhumannet}
\end{table}

\begin{table*}[h]
    \centering
    \scriptsize
    \setlength{\tabcolsep}{5pt}
    \resizebox{0.8\linewidth}{!}{
    \begin{tabular}{l|cccc|cccc}
       Exp.  & JointDiff & $\mathcal{G}_\text{refined}$ & Body parts & $\mathbf{p}_\mathrm{gt}$ & PSNR $\uparrow$ & SSIM $\uparrow$ & LPIPS $\downarrow$ & FID$\downarrow$ \\
       \hline 
        Coarse 3DGS $\mathcal{G}_\mathrm{coarse}$ & \cmark & \xmark & F & \xmark & 20.84 & 0.8789 & 0.1296 & 31.9 \\
        Direct Composition & \cmark & \cmark & U,L,H & \xmark & 20.41 & 0.8700 & 0.1350 & 36.2 \\
        Learnable Composition & \cmark & \cmark & F,U,L,H & \xmark & 20.87 & 0.8788 & 0.1270 & 28.0 \\ 
        No Joint Diffusion & \xmark & \cmark & F,U,L,H & \xmark & 20.79 & 0.8762 & 0.1283 & 27.6 \\ 
        Additional body part & \cmark & \cmark & 
        F,U,M,L,H & \xmark & 21.43 & 0.8922 & 0.1104 & 27.6 \\
        Ours & \cmark & \cmark & F,U,L,H & \xmark & 21.46 & 0.8925 & 0.1087 & 27.3 \\
        Ours + GT Pose Condition & \cmark & \cmark & F,U,L,H & \cmark & \textbf{23.00} & \textbf{0.9028} & \textbf{0.1086} & \textbf{27.0} 
    \end{tabular}
    }
    \vspace{-1.0em}
    \caption{\textbf{Ablation study.} Without ground-truth pose, our full method achieves the best scores compared to the ablation baselines, showcasing the effectiveness of joint diffusion (JointDiff), compositional 3DGS with local refinement ($\mathcal{G}_\mathrm{refined}$), and the selection of body parts (F: fullbody, U: upper, L: lower, H: head, M: middle). Using ground-truth pose ($\mathbf{p}_\mathrm{gt}$) with our pose-conditioned model can further improve the alignment and provide better results.}
    \vspace{-1.0em}
    \label{tab:ablation-study}
\end{table*}

\textbf{Avatar Reposing.} For avatar reposing evaluation, we sample one input pose $\mathbf{p}_\mathrm{in}$ and six target novel poses $\mathbf{p}_\mathrm{target}$ from the video sequence for each unseen subject in the MVHumanNet dataset. Our method takes a single input image $\mathbf{x}_\mathrm{in}$ and pose conditions $\mathbf{p}_\mathrm{in}, \mathbf{p}_\mathrm{target}$ as inputs, directly synthesizes the avatar in the target poses using the Pose-Conditioned Joint 3D Diffusion. We compare our approach with SOTA mesh-based methods SiTH~\cite{ho2024sith} and SIFU~\cite{zhang2023sifu} using the same inputs, which repose characters into target poses using linear blend skinning and the SMPL-X body model. As an additional baseline, we also evaluate results from directly deforming the input pose reconstructed 3DGS avatar into target poses by SMPL blending weights. In particular, other 3DGS-based methods, such as LGM~\cite{tang2025lgm} and Human3Diffusion~\cite{xue2024human3diffusion} are excluded from this evaluation because they do not have aligned body models to support reposing of their reconstructed avatars. As shown in \cref{tab:nps-mvhumannet}, \mn significantly outperforms competing methods across all metrics. \cref{fig:exp-comp-nps} illustrates that our pose-conditioned model generalizes effectively to challenging input and target poses, benefiting from the diverse motions present in multi-view video datasets. Notably, \mn excels at synthesizing realistic cloth deformations in target poses, while other methods struggle due to limitations of SMPL-based deformation and the fixed topology of mesh-based reconstruction methods.

\begin{figure}[h]
\begin{center}
 \centering
 \includegraphics[width=1.0\linewidth]{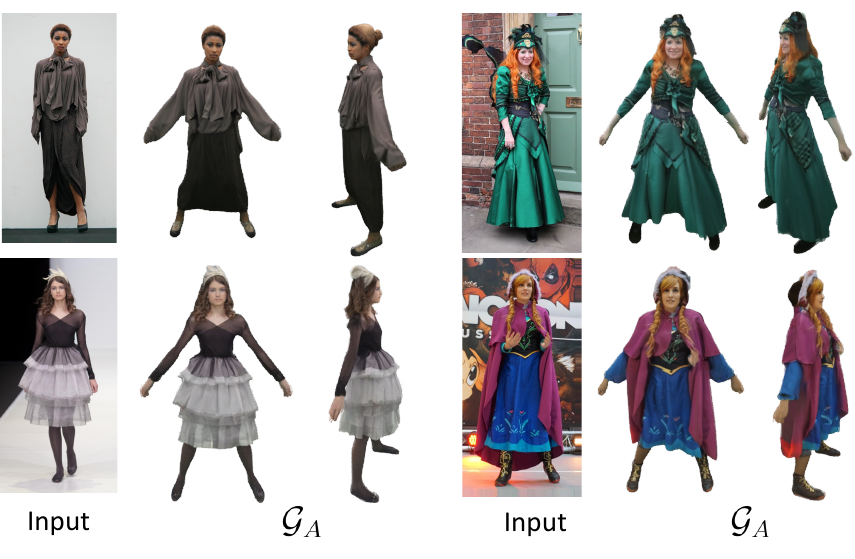}
 \vspace{-2.0em}
 \caption{\textbf{Reposing avatars with challenging garments.} }
 \vspace{-2.0em}
 \label{fig:repose-coarse}
\end{center}
\end{figure}

Additionally, in \autoref{fig:repose-coarse}, we show results of reposing SHHQ\cite{fu2022styleganhuman} characters with complex loose clothing to standard poses. Our reposing model successfully generalize to these OOD garments with realistic deformation effects.

\vspace{1mm}
\noindent\textbf{Avatar Animation.} \cref{fig:exp-animation} showcases the animation results of \mn using the animatable avatar from Avatar Reposing with a standard pose condition. Although the model is not directly trained with standard pose data, it learns to generalize to the standard poses with the help of the diverse distribution of poses in MVHumanNet\cite{xiong2024mvhumannet}.  

\begin{figure}[h]
\begin{center}
 \centering
 \includegraphics[width=1.0\linewidth]{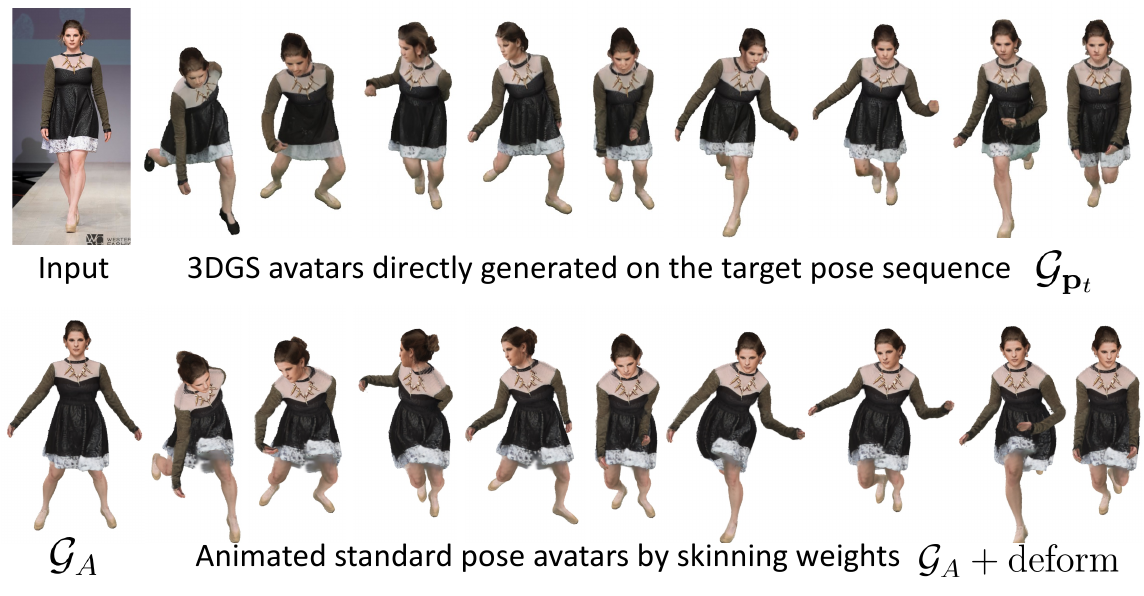}
 \vspace{-2.0em}
 \caption{\textbf{Comparison of direct avatar reposing and standard posed avatar with skinning weight animation.} }
 \vspace{-2.0em}
 \label{fig:comp-animation}
\end{center}
\end{figure}

\noindent\textbf{Avatar Reposing vs. LBS-based Animation}
As \mn supporting two modes to synthesize novel pose avatars, \autoref{fig:comp-animation} compares the performance of these two modes. Here we analysis by comparing their pros and cons. 

\noindent\textit{Mode 1: Direct Avatar Reposing} - This mode directly generates reposed Gaussians for a target pose. Pros: (1) Captures pose-dependent effects for non-rigid clothing, (2) More realistic deformation of loose clothing, (3) No need for rigging. Cons: More computationally expensive and less temporal coherent.

\noindent\textit{Mode 2: SMPL-based LBS Animation} - This mode first reconstructs a standard pose avatar, then applies SMPL-based skinning weights for motion deformation. Pros: (1) Enables real-time rendering, (2) Better temporal consistency. Cons: Limited loose clothing deformation.

\begin{figure}[h]
\begin{center}
 \centering
 \includegraphics[width=1.0\linewidth]{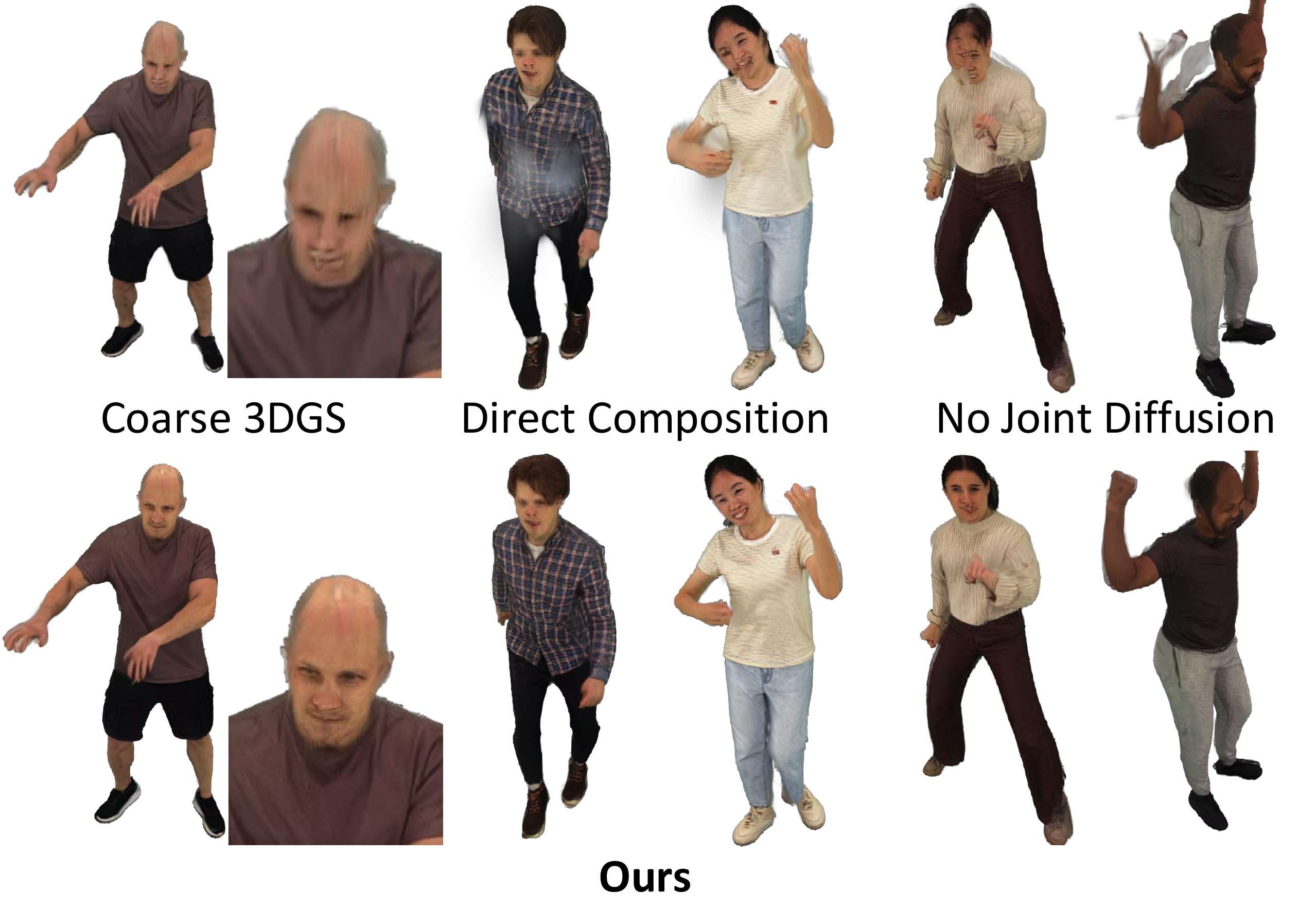}
 \vspace{-2.0em}
 \caption{\textbf{Comparison of our method and ablation variants.} }
 \vspace{-2.0em}
 \label{fig:ablation}
\end{center}
\end{figure}

\subsection{Ablation Study}

To evaluate the effectiveness of our  design choices, we conduct various ablation studies on avatar reconstruction using the CustomHumans dataset. \cref{tab:ablation-study} and \cref{fig:ablation} compare variants of our method, focusing on rendering quality.

\vspace{1mm}
\noindent\textbf{Coarse 3DGS} $\mathcal{G}_\mathrm{coarse}$ uses only the generated coarse avatar without refinement, failing to capture fine details, particularly in facial regions. Our full method achieves better FID while maintaining comparable PSNR, SSIM, and LPIPS scores, demonstrating that compositional refinement improves details without sacrificing accuracy.

\vspace{1mm}
\noindent\textbf{Composition Strategy.} We compare our visibility-aware approach with: (1) \textbf{Direct Composition}, which ensembles all local 3DGS without filtering unreliable splats, yet this variant results in significant artifacts; (2) \textbf{Learnable Composition}, which uses a network with self-attention between parts to predict the holistic avatar. Despite showing slight improvement, this variant still encounters artifacts and requires more computation. This demonstrates the importance and effectiveness of our visibility-aware 3DGS composition.

\vspace{1mm}
\noindent\textbf{Body Part Selection.} To evaluate the design of body part selection, we compare with variants that use an additional body part in the middle of the body for local refinement and 3D composition. This comparison demonstrates that using 4 parts~(fullbody, upper, lower and head) is a good balance between performance and efficiency.

\vspace{1mm}
\noindent\textbf{No Joint Diffusion} is a variant that applies the 3DGS generator only to multiview images from the last diffusion step. Results show that it leads to view inconsistencies and performance drops, confirming the importance of 3D joint diffusion for consistent avatar generation.

\vspace{1mm}
\noindent\textbf{GT Pose Condition} shows that using ground-truth SMPL annotations significantly improves reconstruction quality through better pose alignment, indicating potential for further improvement.

\section{Discussion and Limitations}
\label{sec:Conclusion}

In this paper, we introduced \mn, a novel framework for generating highly-detailed and animatable 3DGS avatars from a single input image. Our approach integrates 3DGS reconstruction within the multi-view diffusion process, ensuring 3D-consistent generation of multiview images as well as 3DGS avatars in both input and novel poses. Furthermore, our visibility-aware compositional 3DGS refinement module significantly enhances the appearance details of the avatars and seemlessly integrates local and global body parts into a coherent 3DGS avatar. Extensive experiments on public benchmarks and in-the-wild images showed that \mn substantially outperforms state-of-the-art methods in both novel view synthesis and novel pose synthesis tasks.

Despite these advancements, some limitations of our method warrant further exploration. The local refinement strategy may encounter difficulties with occluded or poorly covered regions, particularly around hands and arms, leading to artifacts and limiting fine-grained animation in these areas. Additionally, while our model can generate avatars in an animation-friendly standard pose, the animation capability still relies on the alignment of the SMPL body models and their skinning weights, which poses challenges in detailed animation such as facial expressions, hand gestures, and garment deformation. Future work could explore better integration of body models and simulation-based methods, as well as the use of video diffusion model to enhance the animation quality.
{
    \small
    \bibliographystyle{ieeenat_fullname}
    \bibliography{main}
}
\clearpage
\appendix

\section{Implementation Details}

\begin{figure*}[ht]
\begin{center}
 \centering
 \includegraphics[width=1.0\textwidth]{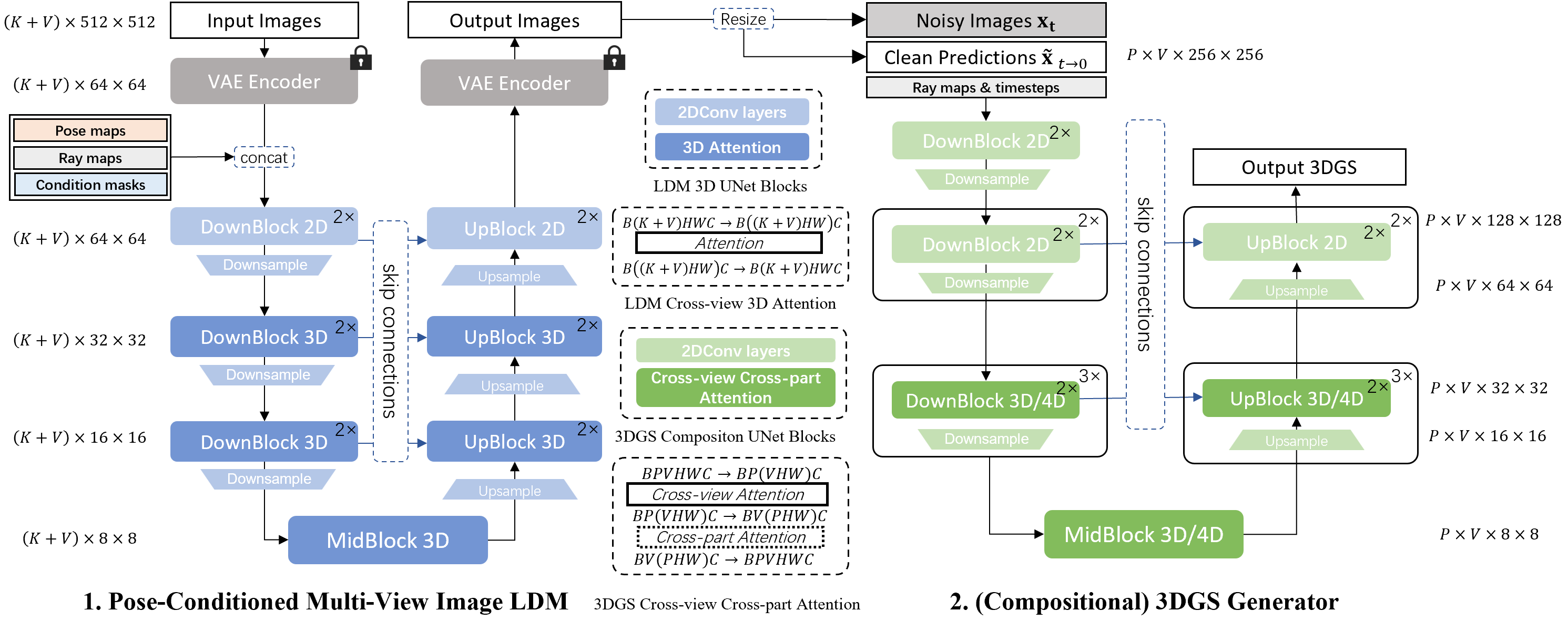}
 \vspace{-2em} 
  \caption{\textbf{Network Architectures} of (1) Pose-Conditioned Multi-View LDM Model and (2) Compositional 3DGS Generator.}
 \label{fig:supp:architecture}
\end{center}
\end{figure*}

\paragraph{Network Structure.} In \cref{fig:supp:architecture}, we illustrate the architecture of our Pose-Conditioned Multi-View Image LDM model, along with the 3DGS generators $\mathbf{G}$ and $\mathbf{G}_\mathrm{comp}$. For the LDM model, following \cite{gao2024cat3d}, we enable 3D cross-view attention only in layers with a feature map resolution of $\le 32\times32$. We also add extra input channels to the latent maps for camera ray maps, condition masks, and semantic pose maps. For $\mathbf{G}$, we adopt the architecture of the pre-trained LGM-big model \cite{tang2025lgm} and include additional input channels for noisy images $\mathbf{x}_t$. 

Additional, as an ablation mentioned at \autoref{tab:ablation-study}, we hvae tried training a compositional 3DGS generator $\mathbf{G}_\mathrm{comp}$ for Learnable Composition. Based on the LGM network, we insert an additional cross-part self-attention layer after each original cross-view self-attention layer in the LGM network. Note that the output image resolution of our LDM model is $512\times 512$, which is then downsampled to $256\times 256$, the input resolution for the 3DGS generator $\mathbf{G}$.

\paragraph{Ray Map Embedding.} We use different methods to embed ray map information for the image LDM model and the 3DGS generators $\mathbf{G}$ and $\mathbf{G}_\mathrm{comp}$. For the 3DGS generators, to effectively utilize the pretrained weights of LGM, we scale the entire scene to ensure a camera distance of $r=1.5$ meters and use Plücker ray embeddings as described in Eq. 4 of the main text.

For the LDM model, we employ sinusoidal positional embeddings \cite{vaswani2017attention} to encode ray origins and directions, providing rich information about 3D locations across different cropping scales:

\begin{equation}
\mathcal{R}_\mathrm{LDM}(i, j) = \mathrm{PE}(\mathbf{o}(i,j), \mathbf{d}(i,j))
\label{equ:ldm-pos-embed}
\end{equation}
where $\mathrm{PE}$ is the sinusoidal positional encoding function, with the number of octaves $N_\mathrm{octaves}$ set to 8.

\paragraph{View Sampling.} Since our training data consists of multi-camera video captures in a 3D scene, the avatar is not always positioned at a standard location. We use 2D joint locations and foreground mask areas to crop global and local training views, resizing them to a resolution of $512\times 512$. In \cref{tab:view-sampling}, we list the OpenPose joints used to determine the cropping centers and relative size ratios of the local crops. During inference, after obtaining coarse reconstruction results with global views, we render $N_v=20$ views to estimate 3D joints using EasyMocap \cite{easymocap}, which helps sample local views for our compositional 3DGS refinement.

\begin{table}[ht]
    \centering
    \resizebox{\linewidth}{!}{
    \begin{tabular}{c|cccc}
         Parts & Full body & Upper Body & Lower Body & Head \\
         \hline
         Joints & Pelvis & Neck & Left Ankle, Right Ankle & Left Ear, Right Ear \\
         Scale & 1.0 & 0.5 & 0.5 & 0.25 
    \end{tabular}
    }
    \vspace{-2mm}
    \caption{Body part sampling details.}
    \label{tab:view-sampling}
\end{table}

\paragraph{Training Schedule.} We initialize our LDM model with the official weights of \texttt{stable-diffusion-v1-5}\footnote{\url{https://huggingface.co/stable-diffusion-v1-5/stable-diffusion-v1-5}}~\cite{Rombach2022StableDiffusion} and our 3DGS generator $\mathbf{G}$ with LGM-big\footnote{\url{https://huggingface.co/ashawkey/LGM/resolve/main/model_fp16_fixrot.safetensors}}~\cite{tang2025lgm}.

For training the LDM model weights $\mathbf{\theta}$, the model first learns to predict $K=3$ canonical views from one input view ($V=1$) without pose conditioning. We fine-tune the model on predicting global full-body views for $20,000$ iterations, followed by fine-tuning on all $N_p+1=4$ global and local view for another $30,000$ iterations to obtain $\mathbf{\theta}_\mathrm{no\_pose}$. Finally, we fine-tune the pose-conditioned model weights $\mathbf{\theta}_\mathrm{novel\_pose}$ from $\mathbf{\theta}_\mathrm{no\_pose}$. This model learns to predict $K=4$ canonical views of a novel pose avatar from $V=1$ input views sampled from different frames in the same video sequence. The novel pose synthesis model is fine-tuned for $1,0000$ iterations using all $N_p+1=4$ global and local views.

For training the 3DGS generator model $\mathbf{G}$, we first fine-tune it from pre-trained weights using clean full-body images in MVHumanNet~\cite{xiong2024mvhumannet} for $2,000$ iterations to adapt it for human reconstruction. Then, we randomly sample diffusion timesteps to train with both noisy inputs $\mathbf{x}_t$ and clean inputs $\mathbf{x}_0$ for $20,000$ iterations. The 3DGS model $\mathbf{G}$ is also fine-tuned on local views for an additional $20,000$ iterations. We use $N_\mathrm{ref}=12$ reference views of each part to supervise the predicted 3DGS.

All training processes are conducted on 16 NVIDIA A100 80GB GPUs, with a total batch size of $n_\mathrm{batch}=128$ and a learning rate of $\eta=5\times 10^{-5}$.

\paragraph{Training Losses.} The training losses for the pose-conditioned LDM and the 3DGS generator are as follows:
\begin{align}
    \mathcal{L}_\mathrm{LDM} &= \mathcal{L}_\mathrm{MSE}(\epsilon,\epsilon_\theta) \\
    \mathcal{L}_\mathbf{G} &= \mathcal{L}_\mathrm{recon} + \lambda_\mathrm{reg}\mathcal{L}_\mathrm{reg} \\
    \begin{split}
    \mathcal{L}_\mathrm{recon} &= \lambda_\mathrm{MSE}\mathcal{L}_\mathrm{MSE}(\hat{\mathbf{x}}_\mathrm{novel}^{t\rightarrow 0}, \mathbf{x}_\mathrm{novel})\\ 
    & + \lambda_\mathrm{LPIPS}\mathcal{L}_\mathrm{LPIPS}(\hat{\mathbf{x}}_\mathrm{novel}^{t\rightarrow 0}, \mathbf{x}_\mathrm{novel})
    \end{split}
\end{align}
where the training loss of LDM, denoted as $\mathcal{L}_\mathrm{LDM}$, is the MSE loss of the predicted latent noise. The training loss of $\mathbf{G}$ consists of rendering reconstruction loss computed using MSE and LPIPS. Following \cite{xue2024human3diffusion}, we also incorporate the 3DGS regularization loss from \cite{Huang2DGS2024,Yu2024GOF} to enhance surface quality.

\paragraph{Inference.} This section details the inference pipeline of avatar reconstruction and avatar reposing our method. In both settings, we perform 3D joint diffusion on global views only when \(t \in (500, 900]\) to maintain the stability of the diffusion process. The earlier steps focus on pure 2D diffusion to generate more detailed appearances. During image-to-image local refinement, we utilize SDEdit \cite{meng2021sdedit} with a strength of \(s=0.5\), meaning that denoising begins at \(t=500\) and 3D joint diffusion is performed when \(t \in (350, 500]\).

\section{Evaluation Settings}

\paragraph{Baseline Models.} Our baseline methods, including Human3Diffusion \cite{xue2024human3diffusion}, LGM \cite{tang2025lgm}, SiTH \cite{ho2024sith}, and SIFU \cite{zhang2023sifu}, have been trained on various 3D mesh datasets \cite{Deitke2023Objaverse, tao2021function4d, ho2023custom}. In this work, our aim is to demonstrate the advantages of training models on both mesh datasets and video datasets for better pose generalization and the synthesis of novel pose characters. We utilize their official weights for comparison. We also note that some models (e.g. \cite{xue2024human3diffusion}) rely on private data or synthesized meshes for training.

\paragraph{Avatar Reconstruction.} We selected front views of the mesh avatar as input views, rendered by horizontal perspective cameras for a fair and realistic comparison. The results of the quantitative evaluation are rendered at a resolution of $1024\times 1024$ using $20$ perspective cameras.

\paragraph{Avatar Reposing.} For SiTH \cite{ho2024sith} and SIFU \cite{zhang2023sifu}, we deform their avatars to the target pose and align the avatar meshes with the ground-truth SMPL meshes to render images for evaluation.

\end{document}